\documentclass[10pt,twocolumn]{article}
\usepackage{simpleConference}
\usepackage{times}
\usepackage{graphicx}
\usepackage{amssymb}
\usepackage{url,hyperref}
\usepackage{graphicx}
\usepackage[caption=false,font=footnotesize]{subfig}

\begin{document}

\title{Classifying Mammographic Breast Density by Residual Learning}

\author{Jingxu Xu
\\
Shenzhen University, Shenzhen, China \\
\\
Cheng Li\\
Shenzhen Institutes of Advanced Technology, Chinese Academy of Sciences, Shenzhen, China \\
\\
Yongjin Zhou\\
Shenzhen University, Shenzhen, China \\
\\
Lisha Mou\\
Shenzhen Second People¡¯s Hospital, the First Affiliated Hospital of Shenzhen University, Shenzhen, China \\
\\
Hairong Zheng,Shanshan Wang*\\
Shenzhen Institutes of Advanced Technology, Chinese Academy of Sciences, Shenzhen, China \\
\\
\today
\\
\\
sophiasswang@hotmail.com; ss.wang@siat.ac.cn  \\
}

\maketitle
\thispagestyle{empty}

\begin{abstract}
Mammographic breast density, a parameter used to describe the proportion of breast tissue fibrosis, is widely adopted as an evaluation characteristic of the likelihood of breast cancer incidence. Existing methods of breast density classification either requires steps of manual operations or achieves only moderate classification accuracies due to the limited model capacity. In this study, we present a radiomics approach based on residual learning for the classification of mammographic breast densities. Different from those established approaches, our method possesses several encouraging properties including being almost fully automatic, possessing big model capacity, and having high flexibility. As a result, it can obtain outstanding classification results without the necessity of result compensation using mammographs taken from different views. The proposed method was instantiated with the INbreast dataset and classification accuracies of 92.6\% and 96.8\% were obtained for the four BI-RADS (Breast Imaging and Reporting Data System) category task and the two BI-RADS category task, respectively. Both values are significantly higher than the classification results of the current state-of-the-art methods, including the eight-layer convolutional neural network and the high throughput-derived multilayer visual representations. The superior performances achieved with its encouraging properties indicate that our method has a great potential to be applied as a computer-aided diagnosis tool.
\end{abstract}

\section{Introduction}
Breast cancer is a big heath threat\cite{cancer2017,Bray2004The}, the incidence of which has increased while its death rates have declined in all age groups in the past decades\cite{Bray2004The,Sickles1997Breast}. This favorable trend of mortality reduction could be related to the improvements in the treatment of breast cancer and the widespread adoption of breast cancer screening techniques, especially mammography\cite{Sickles1997Breast}, for early diagnosis.

Mammography is the most common and efficient method for breast cancer screening. Clinical studies reported that compared to mammographic abnormalities (e.g masses, calcification, architectural distortion, asymmetries), the change of breast density is another important indicator of early breast cancer development\cite{Oliver2008A,Oliver2015Breast,Rampun2018Breast}. However, inspection of the generated large quantities of mammographs by radiologists is tedious and subjective, which also suffers from the intra- and inter-radiologists reproducibility problem\cite{Huo2014Mammographic,Berg2000Breast}.

The very first research on the importance of breast density began with Wolfe \textit{et al}., who demonstrated the relationship between mammographic parenchymal patterns and the risk of developing breast cancer\cite{Wolfe1976Risk}. Following this, Boyd \textit{et al}. showed a similar correlation between mammographic densities and breast cancer risks\cite{Boyd1995Quantitative}. Inspired by these discoveries, a number of studies on breast density classification emerged. The American College of Radiology¡¯s (ACR) Breast Imaging Reporting and Data System (BI-RADS) groups breasts into four categories according to the density with BI-RADS I refers to the lowest densities and BI-RADS IV refers to the highest (BI-RADS I: fat breast (0-25\%), BI-RADS II: fat with some fibroglandular tissue (26-50\%), BI-RADS III: heterogeneously dense breast (51-75\%), and BI-RADS IV: extremely dense breast (76-100\%)).
Female with extremely dense breasts (BI-RADS IV) have a 2-6 times higher risk of developing breast cancer than female with fatty breasts (BI-RADS I)\cite{Huo2014Mammographic,Kallenberg2016Unsupervised}. Therefore, breast density plays an important role in the early detection of breast cancer, and there is an urgent need of an automatic system which can accurately classify mammographic breast densities.

Initially, many studies measured breast density by quantifying the gray-level histograms of mammographs\cite{Karssemeijer1998Automated,Martin2006Mammographic,Zhou2000Computerized}. Subsequent studies found that it might be insufficient to classify breasts into the corresponding BI-RADS categories only based on the information of histograms. For example, the study by Oliver \textit{et al}. illustrated that the four different categories are quite similar with regard to both the mean gray-level values and the shapes of the histogram.

To address this issue, researchers turned to applying traditional feature engineering methods to handle the breast density classification task. Bovis \textit{et al}. got a 71.4\% accuracy by using a classifier paradigm where a combination of the Fourier and discrete wavelet transforms was investigated on the first and second-order statistical features\cite{Bovis2002Classification}. Oliver \textit{et al}. extracted morphological and texture features from breast tissue regions which were segmented using a fuzzy c-means clustering technique, and these features were then treated as inputs for the breast density classifier\cite{Oliver2008A}. Jensen \textit{et al}. adopted the same breast tissue segmentation method but extracted the first and second-order statistical features as well as morphological features for the Mammographic Image Analysis Socity (MIAS) dataset\cite{Jensen2010Fuzzy}. These two studies achieved 86.0\% and 91.4\% breast density classification accuracies, respectively. Chen \textit{et al}. evaluated different local features using texture representation algorithms. After that, they modelled mammographic tissue patterns based on the local tissue appearances in mammographs\cite{Chen2011Local}. The work of Indrajeet \textit{et al}. were based on ROIs manually extracted from image. Then, multi-resolution texture descriptors were extracted from 16 sub-band images which were obtained from second level decomposition through wavelet packet transform\cite{Kumar2015Wavelet}. It could be concluded from these studies that the general procedure of breast density classification includes segmenting the breast area, designing and extracting breast density-related features, and inputting these features into different classifiers to predict the density categories. One major drawback of this procedure is that prior expert knowledge of the data and a hand-crafting process are necessary to calculate the quantitative features.

On the other hand, the development of deep learning fields offers a promising solution of using artificial neural networks to automatically extract features for medical image analysis\cite{Litjens2017A,Shen2017Deep,Shin2016Deep,Zhang2018Convolutional}. Convolutional Neural Network (CNN) is one type of these networks that has shown excellent performances in image classification. CNN can learn highly nonlinear relationships between the inputs and outputs without human intervention. A number of studies have applied deep learning to mammographic related tasks, such as lesion detection, benign and malignant masses differentiation, microcalcification recognition, and their combinations\cite{Jin2016Discrimination,Suzuki2016Mass,DBLP:journals/corr/WangKGIB16,Zhang2018Direct,Carneiro2017Automated}. In respect of the breast density classification, Mohamed \textit{et al}. designed an eight-layers CNN to group the mammographs into two categories (scattered density and heterogeneously dense) as a simplification of the complicated four BI-RADS category task\cite{Mohamed2017A}. Similarly, Ahn \textit{et al}. designed a CNN architecture to learn the image characteristics from mammographs and classify the corresponding breasts into dense and fatty tissues\cite{Ahn2017A}. From these pioneer studies, we can find that few studies directly classified mammographs into the four BI-RADS categories. One possible reason is that the model capacity of CNN models applied were limited whose shallow network structures prevent it from obtaining enough meaningful and abstract features for accomplishing the difficult task.

Radiomics is an emerging method in recent years that works by extracting large amounts of advanced quantitative features from medical images and quantifying the predictive or prognosis relationships between images and medical outcomes according to the features\cite{Lambin2012Radiomics,Gillies2016Radiomics}. Nevertheless, the advantages of CNNs have not been fully integrated with the radiomics approach to solve the problems encountered during classifying mammographic breast densities into the four BI-RADS categories. Therefore, in this paper, we propose a CNN-based (residual learning)\cite{He2016Deep} radiomics method for the automatic extraction of high-throughput features from mammographs and the subsequent classification of the breast densities. Specifically, our contributions are threefold.

1. Our work demonstrates the first attempt of applying a deep CNN as a radiomics approach to automatically extract high-throughput, high-level, and high-abstract features from mammographs, which serves as the basis of an accurate classification model of mammographic breast densities.

2.  In addition to the existing situation, where a two-category classification is studied, our proposed method can accurately classify mammographic breast densities strictly following the four BI-RADS categories. Moreover, our network possesses the capacity to learn deep features for accurate BI-RADS category classification from a single mammographic image. Result compensation from different views (such as the craniocaudal view and the mediolateral oblique view) is not required.

3. Our method could be treated as a baseline of mammographic breast density classification for clinical applications. Due to the large capacity of residual CNNs, our method could be easily adapted to new datasets with new experimental parameters through parameter fine-tuning.

The rest of this paper is organized as follows. Section II gives the detailed description of the dataset and an overview of the CNN methods based on residual learning and radiomics. In Section III, the proposed CNN architecture and the training details including parameter settings and implementation details are presented. The experimental results are introduced in Section IV, followed by a discussion and conclusion in Section V and VI.

\section{Materials And Methods}

\subsection{Dataset}

\begin{figure*}[!t]
    \centering
  \subfloat[]{\includegraphics[width=0.25\linewidth]{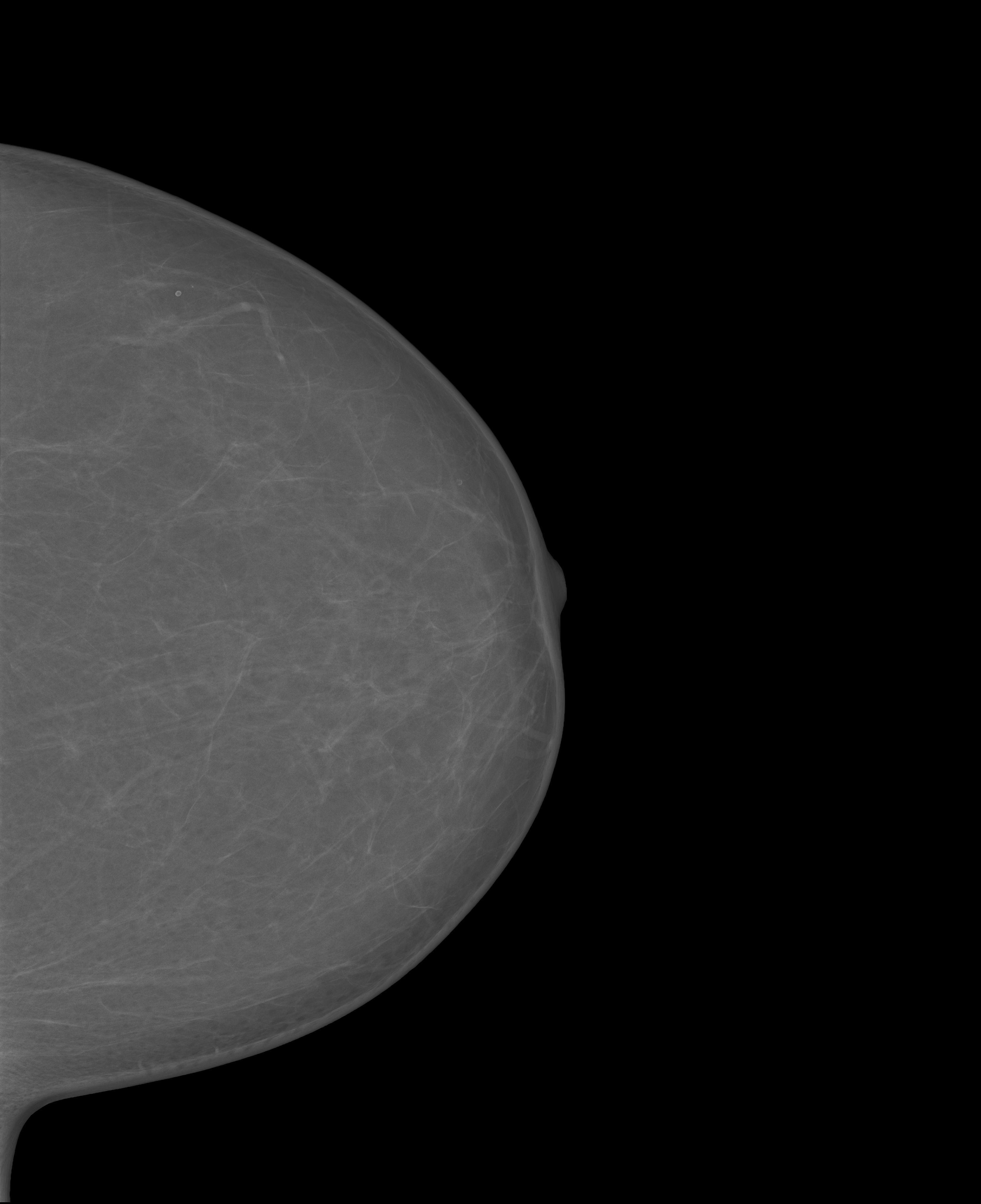}
        \label{SubPeibrain1DAc3}}
  \subfloat[]{\includegraphics[width=0.25\linewidth]{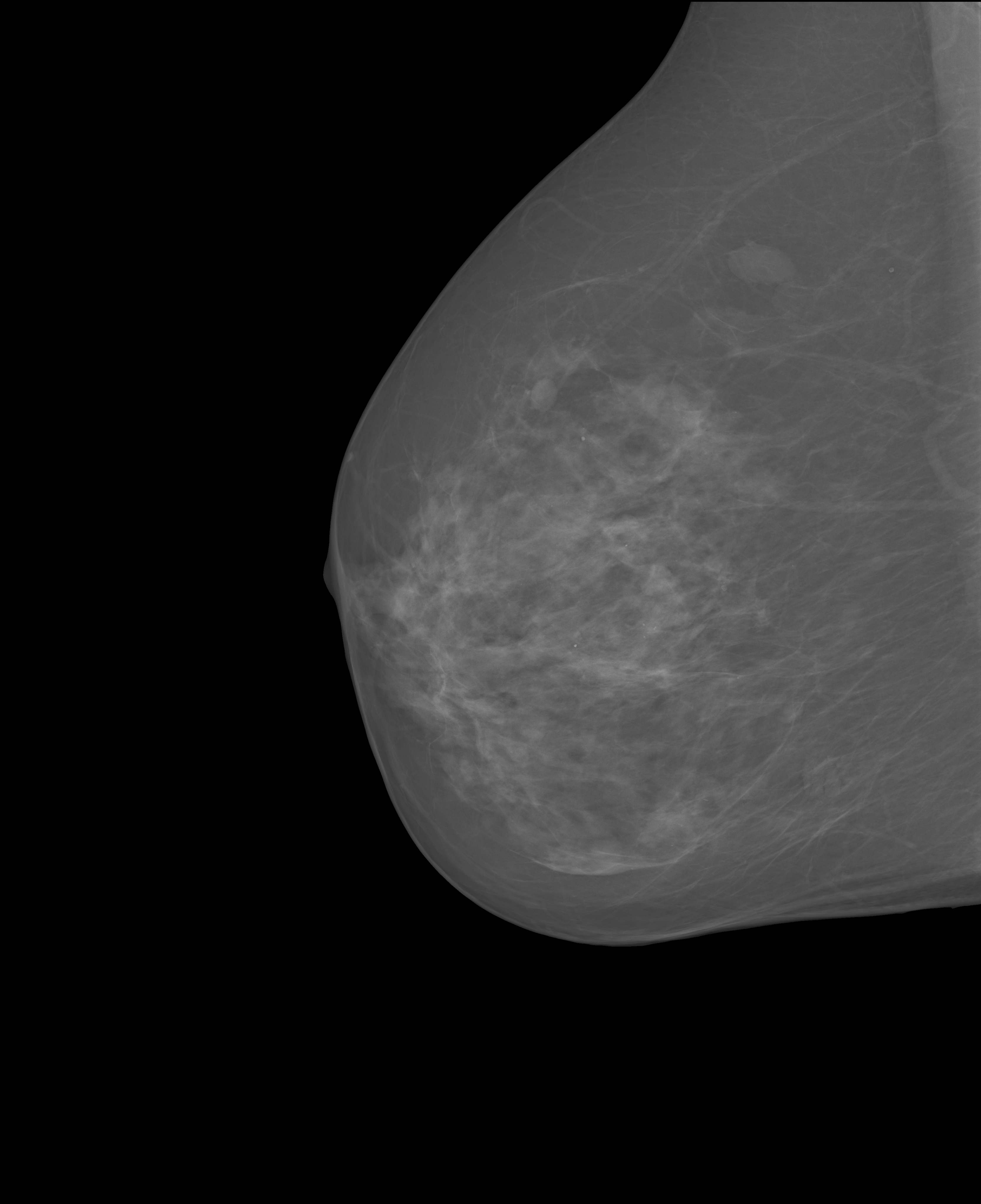}
        \label{GEBrain_2DRandom_RMSE}}
  \subfloat[]{\includegraphics[width=0.25\linewidth]{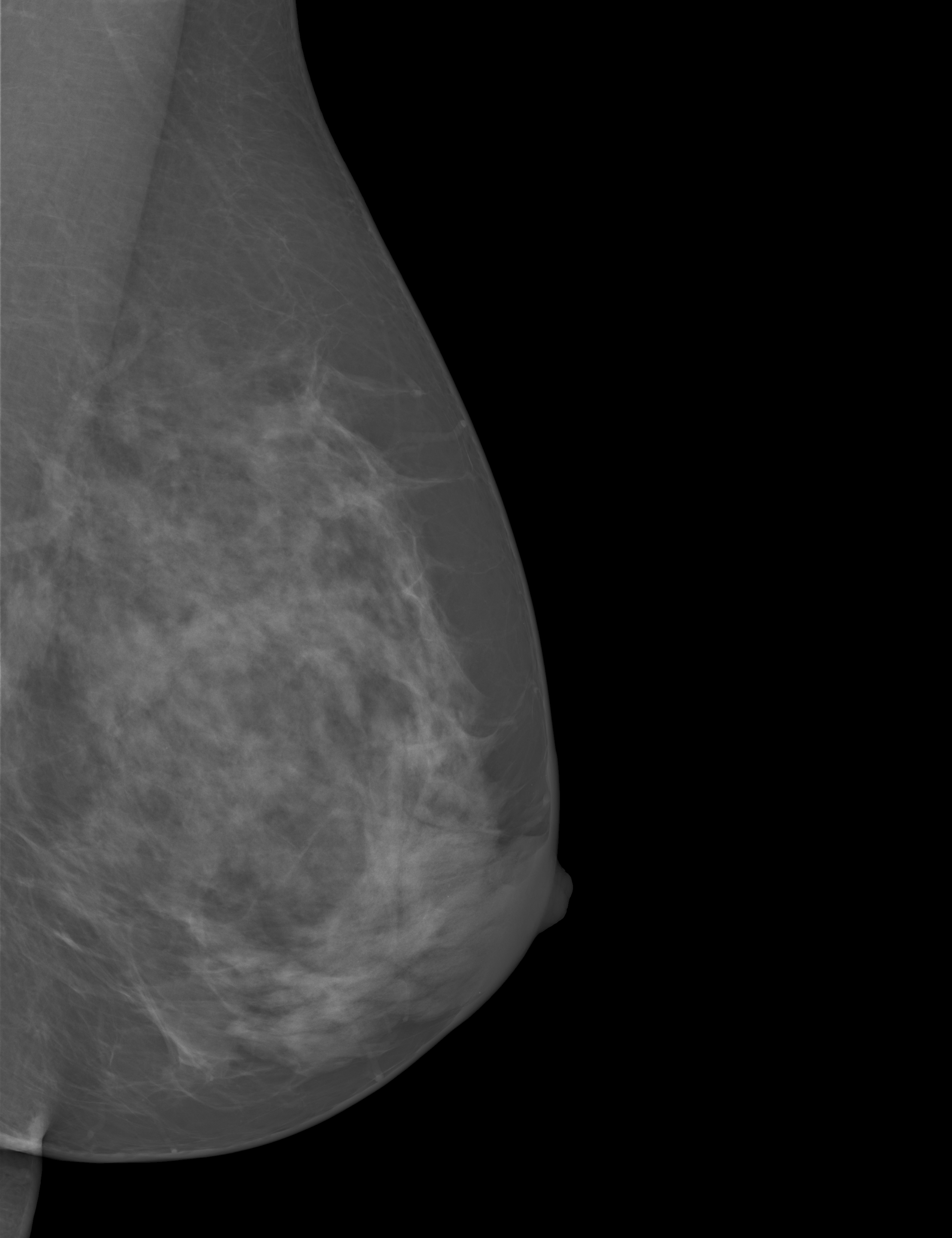}
        \label{GEBrain_2DRandom_RMSE}}
  \subfloat[]{\includegraphics[width=0.25\linewidth]{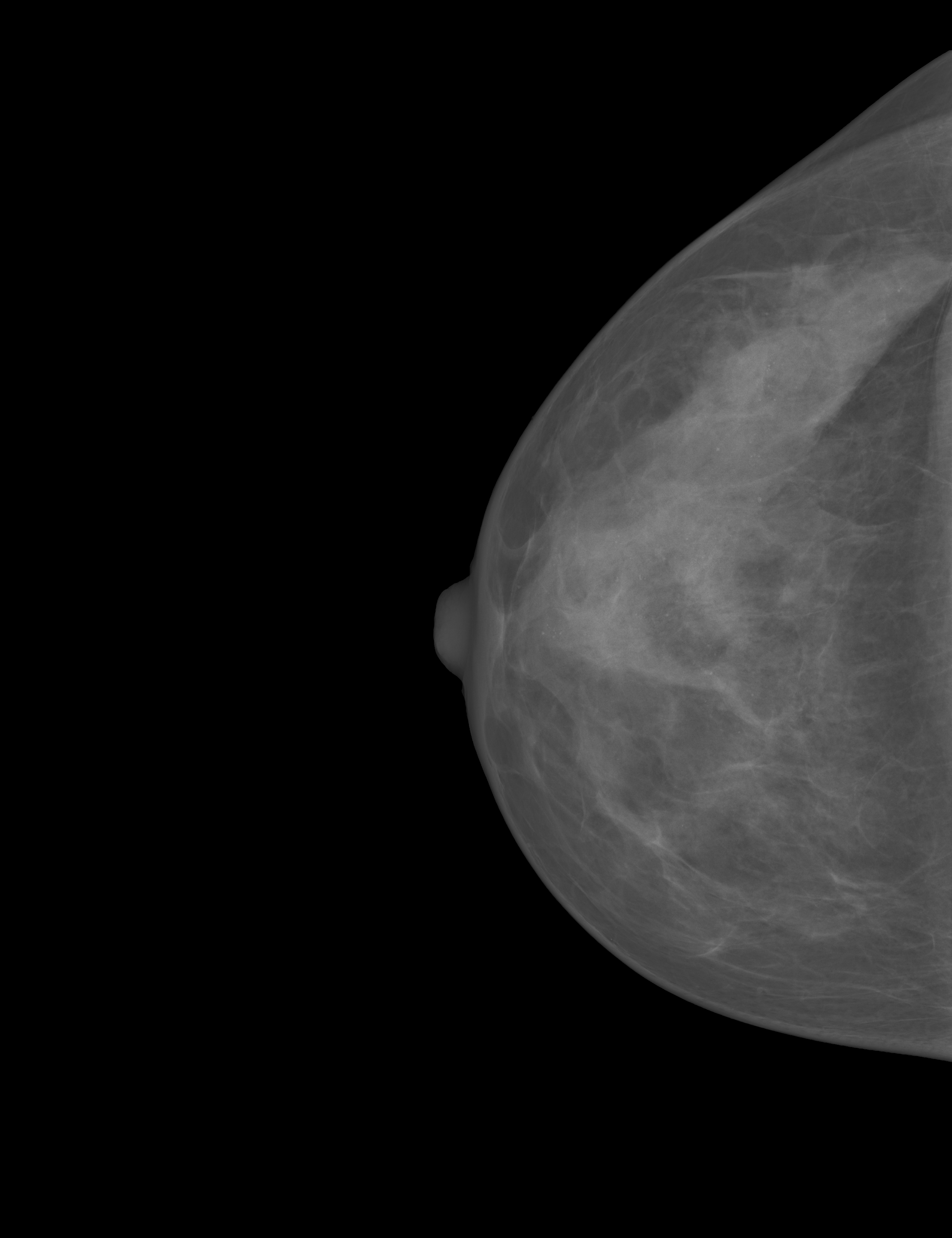}
        \label{GEBrain_2DRandom_RMSE}}
    \caption{\small Example of mammographs of the different BI-RADS categories. (a) BI-RADS I, (b) BI-RADS II, (c) BI-RADS III, and (d) BI-RADS IV.}
    \label{fig1}
\end{figure*}

In this study, we evaluated our methods on the public available dataset, INbreast dataset\cite{Moreira2012INbreast}, which contains 115 cases (410 images). Among the 115 cases, 90 cases are from women with both breasts affected (4 images per case) and 25 cases are from mastectomy patients (2 images per case). Two views for each breast were recorded, a craniocaudal (CC) view, which is a top to bottom view, and a mediolateral oblique (MLO) view, which is a side view. The dataset provides a breast density assessment of each mammograph with the corresponding labels of BI-RADS categories, which makes it suitable for our study. The mammographs were acquired on x-ray films and saved by the standard Digital Imaging and Communications in Medical (DICOM) format. The image matrix has either $3328 \times 4084$ or $2560 \times 3328$ pixels. Among the 409 images(1 missing the label), 136 are classified as BI-RADS I, 146 as BI-RADS II, 99 as BI-RADS III and 28 as class BI-RADS IV (example images of four categories are shown in Fig \ref{fig1}).

\subsection{Data Preprocessing}

As introduced in the dataset section, we have a total number of 409 images. As we all know, training CNN models requires a large number of data and data augmentation is critical step. We also observe a data imbalance between the four BI-RADS categories that needs to be dealt with. We first performed a four-fold rotation augmentation for the BI-RADS IV images. After that, we randomly separated all the images into three groups: 349 for training, 77 for validation and 95 for independent test. At last, to augment the training dataset, we further processed the training and validation sets through rotation by eight random angles, horizontal flip, and vertical flip. Therefore, we have a training dataset of 11168 ($349 \times 8 \times 2 \times 2$) images and a validation dataset of 2464 ($77 \times 8 \times 2 \times 2$).

Another problem that needs to be resolved before the network training is the large size of each image. The original mammography images have $3328 \times 4084$ or $2560 \times 3328$ pixels. To reduce the computational load and memory usage, we need to downsample the original images. To reduce the computational load and memory usage, we downsampled the original images, i.e.  resized the original images to $224 \times224$ pixels.

\subsection{CNN-based residual learning for mammograph classification}

CNNs are a class of deep learning methods that attempt to learn high-level features and attack the computer vision problems, such as classification, detection, segmentation, and so on. Gradient vanishing is a big problem for CNNs with deep layers. Thanks to the invention of the residual network, CNNs can go substantially deeper now than previous. A detailed description of the residual learning block will be presented here. The residual learning was used to solve the degradation problem after stacking a lot of convolution layers. We use $H\left ( X \right )$ to denote the desired nonlinear output feature map of the input feature map x after applying the stacked layers. Now, we let the stacked nonlinear layers fit another mapping of:

\begin{equation}
\label{eq_1}
F\left ( X \right )=H\left ( X \right )-X
\end{equation}
and $H\left ( X \right )$ is recast to
\begin{equation}
\label{eq_2}
F\left ( X \right )+X
\end{equation}

\begin{figure}[!t]
    \centering
  \subfloat{\includegraphics[width=1\linewidth]{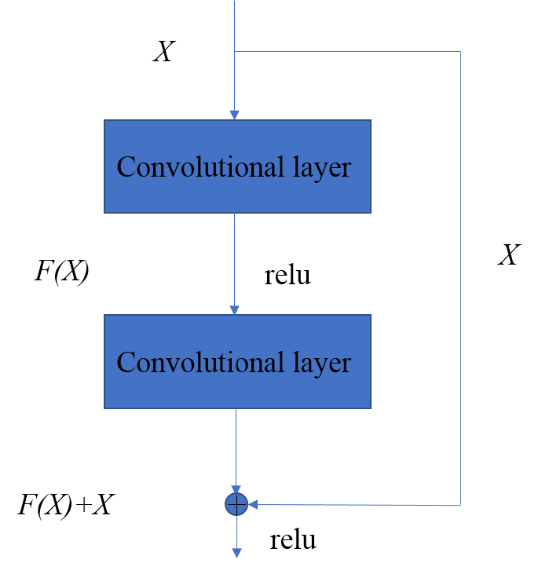}
        \label{SubPeibrain1DAc3}}
    \caption{\small Residual learning block.}
    \label{fig2}
\end{figure}

The formulation of $F\left ( X \right )+X$ can be realized by feedforward CNN with ¡°shortcut connections¡± or ¡°skip connections¡± (Fig\ref{fig2}). In this case, no extra parameters or computational burden are added to the training process. Due to the propagation of gradients through the shortcut connections, it is easier to optimize the residual mapping of $F\left ( X \right )$ than to optimize the original mapping of $H\left ( X \right )$. Therefore, by adding residual learning block, deeper networks could be designed to extract richer information from images for our classification tasks.

\begin{figure*}[!t]
    \centering
  \subfloat{\includegraphics[width=1\linewidth]{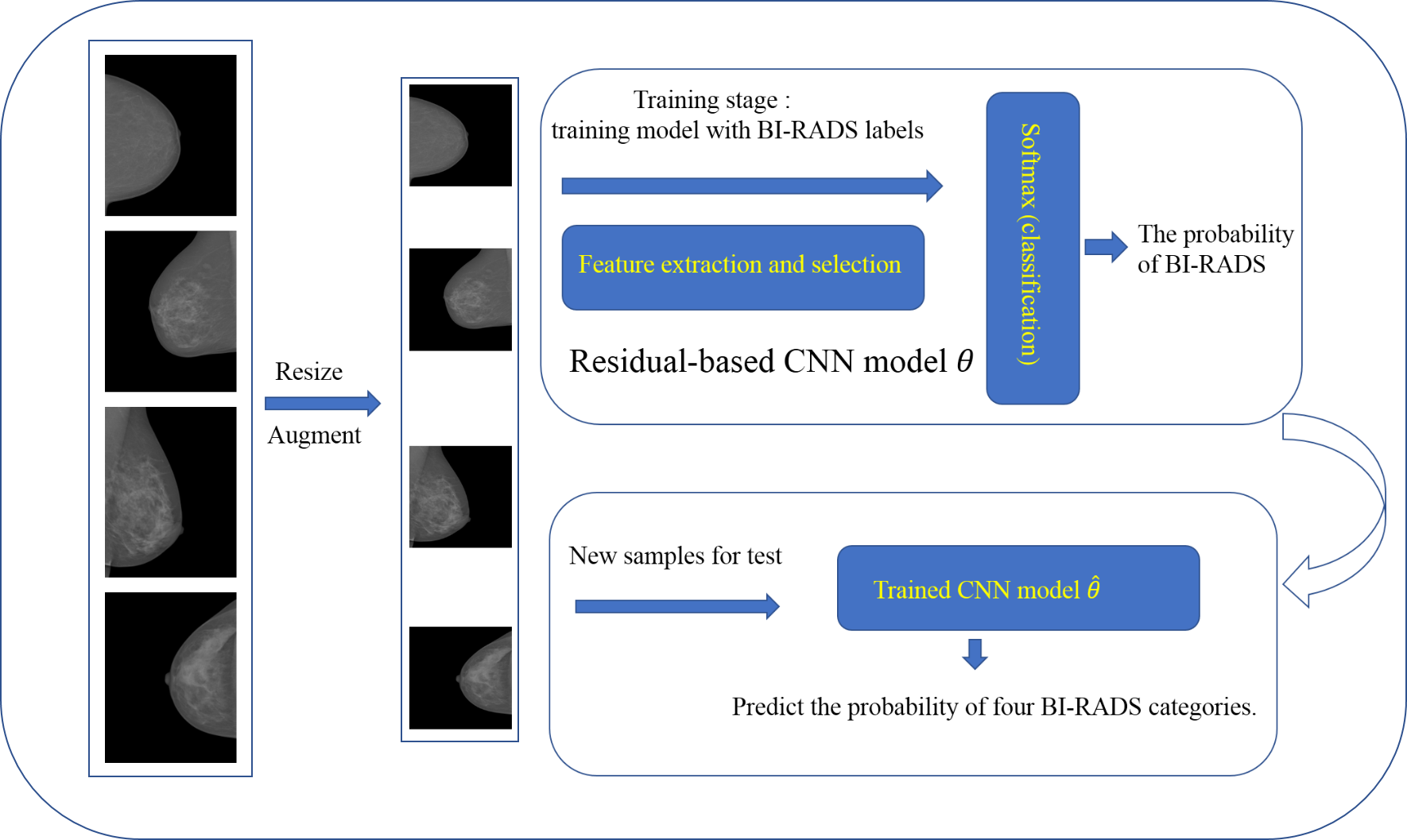}
        \label{SubPeibrain1DAc3}}
    \caption{\small Schematic diagram of residual learning for classification.}
    \label{fig3}
\end{figure*}

Next, we will describe in detail the CNN method used for image classification. After preprocessing, the training, validation, and test datasets went through the training and the test stages respectively as shown in Fig\ref{fig3}. CNNs are trained by feedforward and backpropagation processes. The feedforward process extracts and selects the features and calculates the loss, whereas the backpropagation process optimizes the network parameters by gradient descent of the loss function.

The feedforward process of CNNs could be interpreted by the following steps. Firstly, the images pass through the convolution layers.

\begin{equation}
\label{eq_3}
C_{l} = \sigma_{l}\left ( W_{l}*C_{l-1}+ b_{l}\right )
\end{equation}

Where $l$ denotes the layer number, $\sigma_{1}$ denotes the nonlinear activation (the rectified linear activiation (Relu) was used for this study), $W_{1}$ and $b_{1}$ are weights and bias, * denotes the convolution operation, and $C_{1}$ denotes the feature maps with $C_{0}$ denotes the input. Some convolution layers are followed by downsampling procedure (average pooling layers).

\begin{equation}
\label{eq_4}
C_{l, pool} = averagepooling\left ( C_{l} \right )
\end{equation}

For our classification task, a softmax activation was included after three fully connected layers.

\begin{equation}
\label{eq_5}
C_{L} = softmax\left ( W_{L}C_{L-1}+ b_{L}\right )
\end{equation}

where $softmax\left ( x \right )_{i}=\frac{e^{x_{i}}}{\sum e^{x_{i}} }$ and $C_{L}$ is the output of last layer, So the final prediction from the network could be summarized as
\begin{equation}
\label{eq_6}
Y=C\left ( \theta ,X \right )
\end{equation}
where $\theta$ consists of of all the network parameters to be estimated and optimized, C denotes the overall forward pass network and X means the input and X refers to the input.

On the other hand, the CNN backward process is the backward propagation of loss gradients, which tries to optimize the network parameters ¦È by addressing the following cross-entropy loss minimization problem

\begin{equation}
\label{eq_7}
\hat{\theta }=\mathop{{\rm argmin}}_{{\mathbf \theta}}\left \{ -\sum_{k=1}^{K} \left [ \sum_{i=1}^{I} Y_{i}^{'}\left .log(C\left ( \theta , X \right )  \right ) \right ]\right \}
\end{equation}

where I and K are the total number of classification categories and training samples respectively. $Y_{i}^{'}$ is the manually labelled ground truth provided by the INbreast dataset.

After the training phase, a classification model is obtained with the trained parameters. For new independent samples, we can generate the probability distributions of each case by calculating

\begin{equation}
\label{eq_8}
Y_{test}=C\left ( \hat{\theta },X_{test} \right )
\end{equation}

Then the BI-RADS categories of the mammograph images could be determined accordingly.

\section{Experiments}

\subsection{CNN architecture}

\begin{figure*}[!t]
    \centering
  \subfloat{\includegraphics[width=1\linewidth]{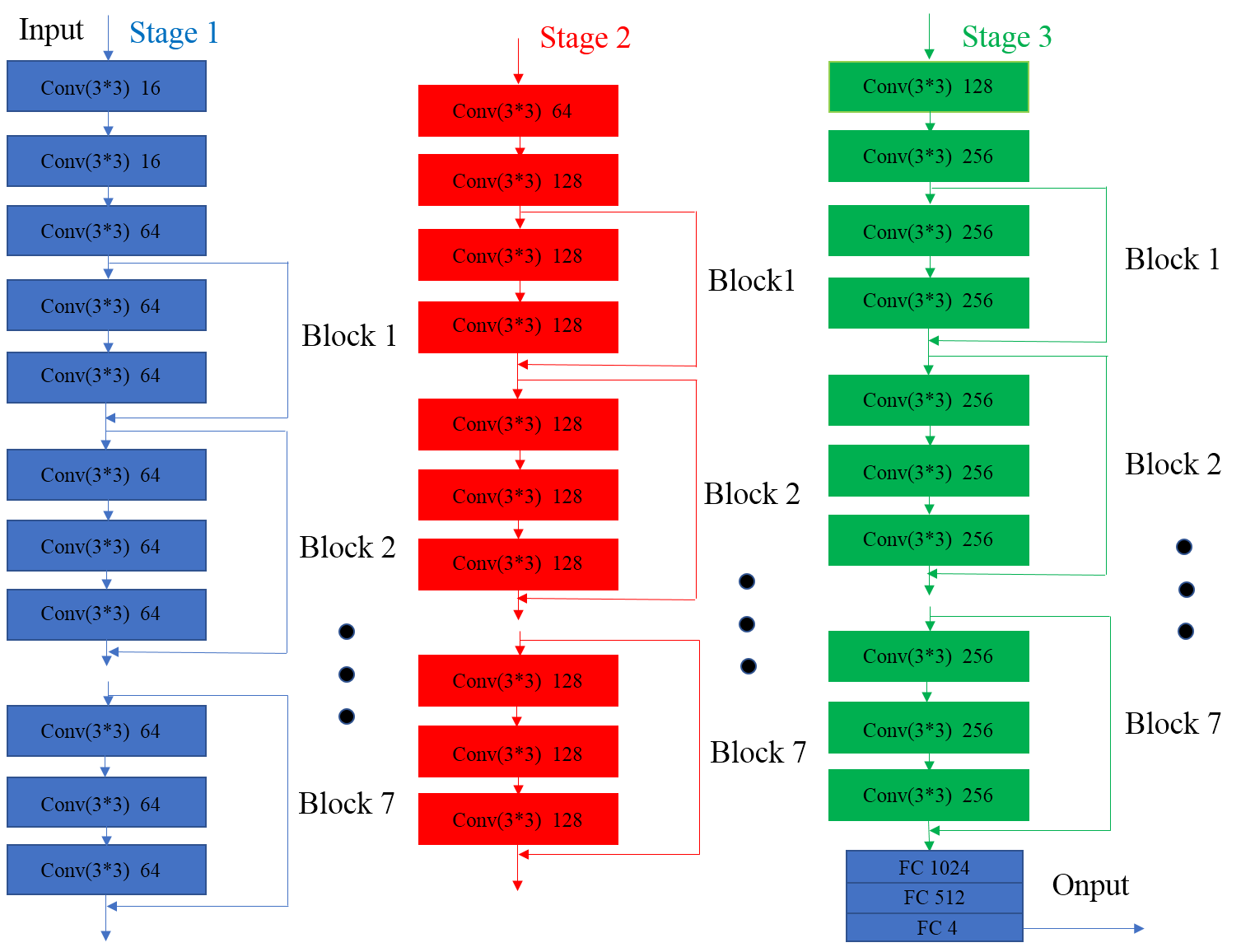}
        \label{SubPeibrain1DAc3}}
    \caption{\small The CNN architecture.}
    \label{fig4}
\end{figure*}

For our classification task, we applied a 70 weight-layer CNN model. As shown in Fig\ref{fig4}, the model could be divided into 3 stages and each stage has 7 residual learning blocks (only 3 residual learning blocks are shown in Fig\ref{fig4}). In total, the model has 70 weight layers (67 convolution layers and 3 fully connected layers) without counting the average pooling and batch normalization layers. All the convolution kernel size was set to $3 \times 3$, and the numbers of convolution kernel for the three residual stages were set to 64, 128, and 256. Moreover, the Relu activation function was adopted following each convolution. The average pooling size and strides were set to $2 \times 2$. For the last three fully connected layers, the number of neurons were respectively set to 1024, 512 and 4. The first two fully connection layers were set with the Relu activation function, while the last one used the softmax instead.

Different layers of the network mean we are extracting different levels of abstract information from the input samples. To test the sensitivity of our classification model to the CNN depth, another two CNN configurations were also evaluated, the 36 and the 48 weight-layer CNN models. These two models also have 3 stages but have fewer convolutional layers in each stage. The different configurations were compared to demonstrate the importance of high-level features on the final classification performance.

\subsection{Parameter settings and implementation details}

We used the Keras (a deep learning framework with tensorflow as backend) to implement our CNN networks for the breast density classification task. Tensorboard was adopted to monitor the entire training process, including the evolution of the accuracy and loss. The network training was implemented on a Dell-7910 workstation equipped with two E5-2640v4 Intel Haswells, a NVIDIA TITAN XP GPU and 64G memory. Adam was used for training, with batch size of 16, maximal number of iterations of 3200 and initial learning rate of 0.0001. Random values drew from the uniform distribution were used for the weight initialization and zero for the bias initialization.

\section{Results}

\subsection{Training convergence property.}

\begin{figure}[!t]
    \centering
  \subfloat{\includegraphics[width=3.5in]{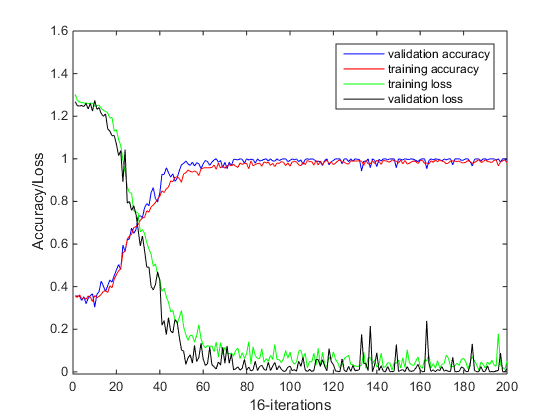}
        \label{SubPeibrain1DAc3}}
    \caption{\small The plot of accuracy and loss in the training stage.}
    \label{fig5}
\end{figure}

Minimizing the cross-entropy loss is the target of the network parameter optimization, and increasing the classification accuracy reflects the improved capability of a classification model to differentiate the different categories. Therefore, to monitor the convergence properties of our network during the training stage, we plotted out the loss as well as the accuracy curves of both the training dataset and the validation dataset with respect to the iterations (Fig. 5). These curves could present the detailed learning procedure of the network. Our loss results fluctuated stably around zero after 80 epochs (Fig\ref{fig5}), which demonstrates that the residual network training converged gradually. The small fluctuations might be caused by the differences between the samples. Similar phenomena could be observed from the accuracy curves with both training and validation curves showed small fluctuations around the accuracy value of 1 after 80 epochs (Fig\ref{fig5}). These results prove that our network training could converge gradually. Once the network was trained, it could be used to obtain the classification predictions of new independent samples.

%

\subsection{Classification performance of different network configurations}

\begin{table}
\centering
\caption{\small PERFORMANCE WITH DIFFERENT NETWORK CONGFIGURATIONS}
\label{}
\begin{tabular*}{8cm}{cccc}
\hline
\hline
Models & 36L  & 48L  & 70L \\
\hline
BI-RADS I  & 92.00\% & 88.00\% & 96.00\% \\
BI-RADS II  & 88.46\% & 100.00\% & 96.15\% \\
BI-RADS III  & 90.48\% & 71.20\% & 95.24\% \\
BI-RADS IV  & 73.91\% & 73.91\% & 82.61\% \\
ALL(accuracy)  & 86.32\% & 85.26\% & 92.63\% \\
\hline
\hline
\end{tabular*}
\end{table}

As explained in the Materials and Methods section, in order to test whether our network is sensitive to the depth of the residual network, we compared the classification results of our 70 weight-layer CNN model to those of the 36 and 48 weight-layer CNN models. Table I summarizes the classification accuracies of the three different network configurations. It could be found that the 36 and 48 weight-layer networks have similar overall classification accuracies. On the other hand, the 70 weight-layer network has a significant increased accuracy. CNN models with different depths can learn features of different hierarchies. We believe that the 70 weight-layer CNN model learned higher levels of features, which led to the improved performance compared to the 36 and 48 weight-layer models. One phenomenon we need to pay attention to is that all of the three networks showed much lower classification accuracies for the BI-RADS IV category, which might be caused by the data imbalance as we only have 28/410 original images that are classified as BI-RADS IV in the INbreast dataset.

\subsection{Classification according to two categories vs four categories}

\begin{table}
\centering
\caption{\small PERFORMANCE WITH DIFFERENT CLASS}
\label{}
\begin{tabular*}{8cm}{cccc}
\hline
\hline
Models & 36L  & 48L  & 70L \\
\hline
Scattered density  & 94.12\% & 98.04\% & 100.00\% \\
Heterogeneously dense  & 97.73\% & 86.36\% & 95.35\% \\
ALL(accuracy)  & 95.79\% & 92.63\% & 96.84\% \\
\hline
\hline
\end{tabular*}
\end{table}

Many studies simplified the problem from the original four-category classification to a two-category classification case. In clinical applications, it is more challenging for radiologists to classify the breasts into the four BI-RADS categories correctly due to the difficulty of discerning the visual features of breast tissues between the four categories. Therefore, some studies treated BI-RADS I and BI-RADS II as one ¡°scattered density¡± category and BI-RADS III and BI-RADS IV as one ¡°heterogeneously dense¡± category in compliance with the clinical requirements.In this respect, we have made small changes to the original residual network to deal with the problem of dichotomous classification. The results are shown in the Table II. We can observe that all of our three residual network configurations showed good dichotomous classification performances, especially the 70 weight-layer residual network. Compared to the other two networks, the 70 weight-layer residual network reached a significantly higher overall classification accuracy of 96.84\%.

\subsection{Comparison with state-of-the-art methods}

In order to further evaluate the proposed method, we compared it to two reported neural network based methods, the eight-layer convolutional neural network\cite{Mohamed2017A} and the high-throughput-derived multilayer visual representations (V1-like, HT-L2 and HT-L3)\cite{Fonseca2015Automatic}. For the first method, the authors explored an eight-layer CNN to classify breasts between scattered density and heterogeneously dense. We applied some non-technical changes to make it possible to compare with our four-class task. For the second method, feature extractors (V1-like, HT-L2 and HT-L3) were first described by Cox \textit{et al}. and Pinto \textit{et al}. for face recognition\cite{Cox2011Beyond,10.1371/journal.pcbi.1000579}. Fonseca \textit{et al}.  applied and evaluated the performance of these feature extractors on classifying mammographs into the four ACR composition categories\cite{Fonseca2015Automatic}. We have made a comprehensive comparison of the different methods considering both two category and four-category classification problems.

From TABLE III and TABLE IV, it could be concluded that for both the two-category problem and the four-category problem, our proposed method always showed higher classification accuracies. One important reason could be that our network was deeper and could extract more abstract and deeper features, which is very important for the accurate classification of the different BI-RADS categories.

\begin{table*}
\centering
\caption{\small CLASSIFICATION ACCURACIES OF DIFFERENT METHODS FOR TWO-CLASS PROBLEM}
\label{}
\begin{tabular*}{12cm}{cccc}
\hline
\hline
Models/Data & Scattered density  & Heterogeneously dense  & ALL(accuracy)  \\
\hline
V1-like  & 94.13\% & 81.82\% & 88.42\% \\
HT-L2  & 96.08\% & 75.00\% & 86.32\% \\
HT-L3  & 94.13\% & 65.10\% & 81.05\% \\
8-CNN  & 96.08\% & 79.55\% & 88.42\% \\
OUR  & 100.00\% & 95.35\% & 96.84\% \\
\hline
\hline
\end{tabular*}
\end{table*}

\begin{table*}
\centering
\caption{\small CLASSIFICATION ACCURACIES OF DIFFERENT METHODS FOR FOUR-CLASS PROBLEM}
\label{}
\begin{tabular*}{12cm}{cccccc}
\hline
\hline
Models & BI-RADS I  & BI-RADS II  & BI-RADS III  & BI-RADS IV & ALL \\
\hline
V1-like  & 72.00\% & 76.92\% & 90.48\% & 56.62\% & 73.68\% \\
HT-L2  & 68.00\% & 76.92\% & 76.20\% & 60.87\% & 70.53\% \\
HT-L3  & 64.00\% & 76.92\% & 95.24\% & 52.17\% & 71.58\% \\
8-CNN  & 76.00\% & 88.46\% & 85.71\% & 65.23\% & 78.95\% \\
OUR  & 96.00\% & 96.15\% & 95.24\% & 82.61\% & 92.63\% \\
\hline
\hline
\end{tabular*}
\end{table*}

\section{Discussion}

Traditional radiomics methods extract features based on manual observation and operation, including manual design, extraction and selection. Compared to the traditional feature engineering approach, deep convolutional networks with residual learning can automatically extract high-order, high-abstraction, and subtle features from mammograms that are not easily observable to human eyes, which enables accurate discrimination of the four BI-RADS categories. Moreover, by working with the whole original images, the classification model has access to all the image-relevant information and elevated performance could be expected. With the proposed method, an overall accuracy of 92.63\% for the four BI-RADS category classification task and an accuracy of 96.84\% for the two BI-RADS category classification task were obtained. Both were higher than the literature reported values, where only relatively shallow networks were applied.

An exam of breast cancer screening by mammography generally comes with CC and MLO views for a single breast. Multi-view models which make a classification decision by considering the different views have been reported. However, to accommodate the information from different views into the final prediction, different model parameter sets need to be trained accordingly, which leads to significantly increased computational burden and decreased testing speed. On the other hand, our proposed model has already shown excellent performance for the mammographic density classification task without considering the relationships between the different views. Therefore, we could conclude that the large capacity of our model enables the extraction of deep enough features for accurate BI-RADS category classifications of breasts which avoids the necessity of multi-view compensation.

Different imaging systems or experimental settings generate images of different standards. A trained CNN can only properly handle the domain-specific images. Though including different types of images into the training process can help build a more robust CNN model, it is not realistic to collect a dataset which considers all the different possibilities. Thanks to the large capacity of CNNs, our classification model could be easily extended depending on the application situations. If the dataset to be processed is in a similar domain as the original dataset, the trained CNN model could be used directly. However, if the new dataset is in a very different domain from the original dataset, fine-tuning of the trained CNN is required before it could be successfully applied. Compared with training from scratch, fine-tuning of CNNs requires much fewer samples and the training process is significantly faster.

Our residual learning-based CNN model could serve as a baseline for mammographic breast density classification. In the future, we expect to collect more data, especially those of BI-RADS IV category, to train a more powerful CNN model. We also plan to test the fine-tuning performance of the baseline model by using datasets that come from different systems or different experimental settings. Finally, we have already collected a number of clinical samples, we will try our method on these samples to investigate the domain transfer behavior of the established model. We will make our code and trained models publicly available once our manuscript gets accepted to foster the research in the field.

\section{Conclusion}

In this study, we have investigated the use of a radiomics-based method through residual learning for mammographic breast density classification. To the best of our knowledge, it is the first attempt that applies residual learning as a radiomics approach to extract high-throughput features from mammographs and classify the breasts accordingly. The superior classification accuracies achieved demonstrate its feasibility. Another important advantage of the proposed method is that the classification model is trained end-to-end. Sophisticated pre-processing of the mammographic images such as segmentation of the breast tissues is not required. This allows the proposed method to be automatic and almost no human intervention is needed. In addition, our method has an appealing attribute that it is readily extendable to different experimental settings and application situations. All of these encouraging properties make it a good candidate algorithm for CAD systems.

\bibliographystyle{abbrv}
\bibliography{refs}

\begin{thebibliography}{10}

\bibitem{Ahn2017A}
C.~K. Ahn, C.~Heo, H.~Jin, and J.~H. Kim.
\newblock A novel deep learning-based approach to high-accuracy breast density
  estimation in digital mammography.
\newblock In {\em Society of Photo-Optical Instrumentation Engineers}, page
  101342O, 2017.

\bibitem{Berg2000Breast}
W.~A. Berg, C.~Campassi, P.~Langenberg, and M.~J. Sexton.
\newblock Breast imaging reporting and data system: inter- and intraobserver
  variability in feature analysis and final assessment.
\newblock {\em Ajr American Journal of Roentgenology}, 174(6):1769--77, 2000.

\bibitem{Bovis2002Classification}
K.~Bovis and S.~Singh.
\newblock Classification of mammographic breast density using a combined
  classifier paradigm.
\newblock {\em International Workshop on Digital Mammography}, pages 177--180,
  2002.

\bibitem{Boyd1995Quantitative}
N.~F. Boyd, J.~W. Byng, R.~A. Jong, E.~K. Fishell, L.~E. Little, A.~B. Miller,
  G.~A. Lockwood, D.~L. Tritchler, and M.~J. Yaffe.
\newblock Quantitative classification of mammographic densities and breast
  cancer risk: results from the canadian national breast screening study.
\newblock {\em Journal of the National Cancer Institute}, 87(9):670--675, 1995.

\bibitem{Bray2004The}
F.~Bray, P.~Mccarron, and D.~M. Parkin.
\newblock The changing global patterns of female breast cancer incidence and
  mortality.
\newblock {\em Breast Cancer Research}, 6(6):229--39, 2004.

\bibitem{Carneiro2017Automated}
G.~Carneiro, J.~Nascimento, and A.~P. Bradley.
\newblock Automated analysis of unregistered multi-view mammograms with deep
  learning.
\newblock {\em IEEE Transactions on Medical Imaging}, PP(99):1--1, 2017.

\bibitem{Chen2011Local}
Z.~Chen, E.~Denton, and R.~Zwiggelaar.
\newblock Local feature based mammographic tissue pattern modelling and breast
  density classification.
\newblock In {\em 2011 4th international conference on biomedical engineering
  and informatics}, pages 351--355, 2011.

\bibitem{Cox2011Beyond}
D.~Cox and N.~Pinto.
\newblock Beyond simple features: A large-scale feature search approach to
  unconstrained face recognition.
\newblock In {\em IEEE International Conference on Automatic Face \& Gesture
  Recognition and Workshops}, pages 8--15, 2011.

\bibitem{Fonseca2015Automatic}
P.~Fonseca, J.~Ferrer, J.~Pinto, and B.~Castaneda.
\newblock Automatic breast density classification using a convolutional neural
  network architecture search procedure.
\newblock In {\em Medical Imaging 2015: Computer-Aided Diagnosis}, 2015.

\bibitem{He2016Deep}
K.~He, X.~Zhang, S.~Ren, and J.~Sun.
\newblock Deep residual learning for image recognition.
\newblock In {\em IEEE Conference on Computer Vision and Pattern Recognition},
  pages 770--778, 2016.

\bibitem{Huo2014Mammographic}
C.~W. Huo, G.~L. Chew, K.~L. Britt, W.~V. Ingman, M.~A. Henderson, J.~L.
  Hopper, and E.~W. Thompson.
\newblock Mammographic density¡ªa review on the current understanding of its
  association with breast cancer.
\newblock {\em Breast Cancer Research \& Treatment}, 144(3):479--502, 2014.

\bibitem{Jensen2010Fuzzy}
R.~Jensen, Q.~Shen, and R.~Zwiggelaar.
\newblock Fuzzy-rough approaches for mammographic risk analysis.
\newblock {\em Intelligent Data Analysis}, 14(2):225--244, 2010.

\bibitem{Jin2016Discrimination}
C.~Jin, H.~Cai, J.~Wang, L.~Li, W.~Tan, and Y.~Xi.
\newblock Discrimination of breast cancer with microcalcifications on
  mammography by deep learning.
\newblock {\em Scientific Reports}, 6:27327, 2016.

\bibitem{Kallenberg2016Unsupervised}
M.~Kallenberg, K.~Petersen, M.~Nielsen, A.~Ng, P.~Diao, C.~Igel, C.~Vachon,
  K.~Holland, N.~Karssemeijer, and M.~Lillholm.
\newblock Unsupervised deep learning applied to breast density segmentation and
  mammographic risk scoring.
\newblock {\em IEEE Trans Med Imaging}, 35(5):1322--1331, 2016.

\bibitem{Karssemeijer1998Automated}
N.~Karssemeijer.
\newblock Automated classification of parenchymal patterns in mammograms.
\newblock {\em Physics in Medicine \& Biology}, 43(2):365, 1998.

\bibitem{Kumar2015Wavelet}
I.~Kumar, H.~S. Bhadauria, and J.~Virmani.
\newblock Wavelet packet texture descriptors based four-class {BIRADS} breast
  tissue density classification.
\newblock {\em Procedia Computer Science}, 70:76--84, 2015.

\bibitem{Lambin2012Radiomics}
P.~Lambin et~al.
\newblock Radiomics: Extracting more information from medical images using
  advanced feature analysis.
\newblock {\em European Journal of Cancer}, 48(4):441--6, 2012.

\bibitem{Litjens2017A}
G.~Litjens, T.~Kooi, B.~E. Bejnordi, S.~Aaa, F.~Ciompi, M.~Ghafoorian, V.~D.~L.
  Jawm, G.~B. Van, and C.~I. S¨¢nchez.
\newblock A survey on deep learning in medical image analysis.
\newblock {\em Medical Image Analysis}, 42(9):60--88, 2017.

\bibitem{Martin2006Mammographic}
K.~E. Martin, M.~A. Helvie, C.~Zhou, M.~A. Roubidoux, J.~E. Bailey,
  C.~Paramagul, C.~E. Blane, K.~A. Klein, S.~S. Sonnad, and H.~P. Chan.
\newblock Mammographic density measured with quantitative computer-aided
  method: comparison with radiologists' estimates and {BI-RADS} categories.
\newblock {\em Radiology}, 240(3):656--65, 2006.

\bibitem{Mohamed2017A}
A.~A. Mohamed, W.~A. Berg, H.~Peng, Y.~Luo, R.~C. Jankowitz, and S.~Wu.
\newblock A deep learning method for classifying mammographic breast density
  categories.
\newblock {\em Medical Physics}, 45(1), 2017.

\bibitem{Moreira2012INbreast}
I.~C. Moreira, I.~Amaral, I.~Domingues, A.~Cardoso, M.~J. Cardoso, and J.~S.
  Cardoso.
\newblock {IN}breast: toward a full-field digital mammographic database.
\newblock {\em Academic Radiology}, 19(2):236--248, 2012.

\bibitem{Oliver2008A}
A.~Oliver, J.~Freixenet, R.~Marti, J.~Pont, E.~Perez, E.~R.~E. Denton, and
  R.~Zwiggelaar.
\newblock A novel breast tissue density classification methodology.
\newblock {\em IEEE Transactions on Information Technology in Biomedicine A
  Publication of the IEEE Engineering in Medicine \& Biology Society},
  12(1):55, 2008.

\bibitem{Oliver2015Breast}
A.~Oliver, M.~Tortajada, X.~Llad¨®, J.~Freixenet, S.~Ganau, L.~Tortajada,
  M.~Vilagran, M.~Sent¨ªs, and R.~Mart¨ª.
\newblock Breast density analysis using an automatic density segmentation
  algorithm.
\newblock {\em Journal of Digital Imaging}, 28(5):1--9, 2015.

\bibitem{10.1371/journal.pcbi.1000579}
N.~Pinto, D.~Doukhan, J.~J. DiCarlo, and D.~D. Cox.
\newblock A high-throughput screening approach to discovering good forms of
  biologically inspired visual representation.
\newblock {\em PLOS Computational Biology}, 5:1--12, 11 2009.

\bibitem{Rampun2018Breast}
A.~Rampun, B.~Scotney, P.~Morrow, H.~Wang, and J.~Winder.
\newblock Breast density classification using local quinary patterns with
  various neighbourhood topologies.
\newblock 4(1):14, 2018.

\bibitem{Gillies2016Radiomics}
G.~RJ, K.~PE, and H.~H.
\newblock Radiomics: Images are more than pictures, they are data.
\newblock {\em Radiology}, 278(2):563, 2016.

\bibitem{Shen2017Deep}
D.~Shen, G.~Wu, and H.~I. Suk.
\newblock Deep learning in medical image analysis.
\newblock {\em Annual Review of Biomedical Engineering}, 19(1):221--248, 2017.

\bibitem{Shin2016Deep}
H.~C. Shin, H.~R. Roth, M.~Gao, L.~Lu, Z.~Xu, I.~Nogues, J.~Yao, D.~Mollura,
  and R.~M. Summers.
\newblock Deep convolutional neural networks for computer-aided detection:
  {CNN} architectures, dataset characteristics and transfer learning.
\newblock {\em IEEE Transactions on Medical Imaging}, 35(5):1285, 2016.

\bibitem{Sickles1997Breast}
E.~A. Sickles.
\newblock Breast cancer screening outcomes in women ages 40-49: clinical
  experience with service screening using modern mammography.
\newblock {\em Jnci Monographs}, 22(22):99--104, 1997.

\bibitem{cancer2017}
R.~Siegel et~al.
\newblock Cancer facts\&figures 2017.
\newblock American Cancer Society
  Availables:https://www.cancer.org/research/cancer-facts-statistics/all-cancer-facts-figures/cancer-facts-figures-2017.html.

\bibitem{Suzuki2016Mass}
S.~Suzuki, X.~Zhang, N.~Homma, K.~Ichiji, N.~Sugita, Y.~Kawasumi, T.~Ishibashi,
  and M.~Yoshizawa.
\newblock Mass detection using deep convolutional neural network for
  mammographic computer-aided diagnosis.
\newblock In {\em Society of Instrument and Control Engineers of Japan}, pages
  1382--1386, 2016.

\bibitem{DBLP:journals/corr/WangKGIB16}
D.~Wang, A.~Khosla, R.~Gargeya, H.~Irshad, and A.~H. Beck.
\newblock Deep learning for identifying metastatic breast cancer.
\newblock {\em CoRR}, abs/1606.05718, 2016.

\bibitem{Wolfe1976Risk}
J.~N. Wolfe.
\newblock Risk for breast cancer development determined by mammographic
  parenchymal pattern.
\newblock {\em Cancer}, 37(5):2486¨C2492, 1976.

\bibitem{Zhang2018Direct}
L.~Zhang, S.~Jiang, Y.~Zhao, J.~Feng, B.~W. Pogue, and K.~D. Paulsen.
\newblock Direct regularization from co-registered contrast {MRI} improves
  image quality of {MRI}-guided near-infrared spectral tomography of breast
  lesions.
\newblock {\em IEEE Transactions on Medical Imaging}, PP(99):1--1, 2018.

\bibitem{Zhang2018Convolutional}
L.~Zhang, L.~Lu, R.~M. Summers, E.~Kebebew, and J.~Yao.
\newblock Convolutional invasion and expansion networks for tumor growth
  prediction.
\newblock {\em IEEE Transactions on Medical Imaging}, 37(2):638, 2018.

\bibitem{Zhou2000Computerized}
C.~Zhou, H.~P. Chan, N.~Petrick, B.~Sahiner, M.~A. Helvie, M.~A. Roubidoux,
  L.~M. Hadjiiski, and M.~M. Goodsitt.
\newblock Computerized image analysis: estimation of breast density on
  mammograms.
\newblock In {\em Medical Imaging 2000: Image Processing}, pages 1056--1069,
  2000.

\end{thebibliography}
\end{document}